\documentclass[10pt,twocolumn,letterpaper]{article}

\usepackage{iccv}
\usepackage{times}
\usepackage{epsfig}
\usepackage{graphicx}
\usepackage{amsmath}
\usepackage{amssymb}

\usepackage{latexsym}
\usepackage{amsfonts}
\usepackage{mathtools}
\usepackage{amsthm}
\usepackage{mathrsfs}
\usepackage{booktabs}
\usepackage[skip=3pt]{caption}

\usepackage[breaklinks=true,bookmarks=false]{hyperref}

\newcommand{\todo}[1]{}
\newcommand{\note}[1]{}

\newcommand{\simm}{\mathrm{sim}}
\newcommand{\ind}[1]{I_{\{#1\}}}
\DeclareMathOperator*{\argmax}{arg\!\max}

\iccvfinalcopy % *** Uncomment this line for the final submission

\begin{document}

%%%%%%%%% TITLE
\title{Renderers are Good Zero-Shot Representation Learners: Exploring Diffusion Latents for Metric Learning}

\author{Michael Tang\textsuperscript{*\ensuremath\heartsuit}\\
% Department of Computer Science, Princeton University\\
% Institution1 address\\
{\tt\small mwtang@princeton.edu}
% For a paper whose authors are all at the same institution,
% omit the following lines up until the closing ``}''.
% Additional authors and addresses can be added with ``\and'',
% just like the second author.
% To save space, use either the email address or home page, not both
\and
David Shustin\textsuperscript{*\ensuremath\heartsuit}\\
% Department of Computer Science, Princeton University\\
% First line of institution2 address\\
{\tt\small dshustin@princeton.edu}
}

\maketitle
% Remove page # from the first page of camera-ready.
\ificcvfinal\thispagestyle{empty}\fi
\def\thefootnote{*}\footnotetext{Equal contribution}
\def\thefootnote{\ensuremath\heartsuit}\footnotetext{Department of Computer Science, Princeton University}
%%%%%%%%% ABSTRACT
\begin{abstract}
    Can the latent spaces of modern generative neural rendering models serve as representations for 3D-aware discriminative visual understanding tasks? We use retrieval as a proxy for measuring the metric learning properties of the latent spaces of Shap-E \cite{shape}, including capturing view-independence and enabling the aggregation of scene representations from the representations of individual image views, and find that Shap-E representations outperform those of the classical EfficientNet baseline representations zero-shot, and is still competitive when both methods are trained using a contrative loss. These findings give preliminary indication that 3D-based rendering and generative models can yield useful representations for discriminative tasks in our innately 3D-native world. Our code is available at \url{https://github.com/michaelwilliamtang/golden-retriever}.
\end{abstract}

%%%%%%%%% BODY TEXT
\section{Introduction}
Recent advances in neural rendering have enabled 2D novel views and entire 3D models to be generated from as few as a single input 2D view, with incredibly high quality \cite{zero123, nerf, dalle2}. Early work on novel view synthesis, focusing on neural radiance fields, takes a \textit{prior-agnostic} approach, training a feature volume on many views of a single scene at inference time before using an algorithmic rendering equation to sample density and color predictions along rays in the novel view to construct it \cite{nerf}. More recently, both neural radiance field and diffusion-based methods take a \textit{prior-based} approach, leveraging large datasets of 2D and 3D scenes to learn priors, and computing novel views or 3D models of a held-out scene by performing forward passes through the model at inference time \cite{zero123, pixelnerf}. This confers the advantage of being able to infer on less input views, which removes cumbersome inference-time data collection constraints, as well as eliminates the need to undergo multi-hour training procedures at inference time (although recent work has found various optimizations that drastically reduce this training process \cite{melon, instantngp}). From an information theoretic perspective, learning 3D-aware priors on a large dataset captures a large amount of geometric information which can significantly reduce the difficulty of the novel view synthesis task, and more closely resembles the conceptual view synthesis processes of humans. For complex scenes, where it is geometrically impossible to capture all of the details from only a handful of camera views, these priors are incredibly important — this means downstream application users can focus on imaging the most important or notable parts of a scene while letting this latent knowledge fill in the more mundane details (e.g. the bottom of a chair or table). For these reasons, we focus on \textit{prior-based} approaches.

Although the origin of these new methods is in generative tasks, a large number of vision and graphics applications involve discriminative understanding -- for example, robotics applications rely on vision models as world models, a way of capturing camera data that leads to sophisticated, natural understanding of scenes, directly using those visual features as inputs into reinforcement learning and control modules \cite{robotics_cnn}, and visual-question-answering models may require strong 3D-aware priors to answer questions that discuss the geometry of a 2D input image, without any explicit generation outputs \cite{vqa}.

We posit that learning strong priors for generative tasks \textit{also leads to good learned 3D-aware representations}. Specifically, we examine view independence as a proxy for 3D-awareness in the latent representations of a neural rendering model (Shap-E \cite{shape}) and a classical neural 2D image understanding model (EfficientNet \cite{efficientnet}). We investigate this by constructing a 2D image retrieval task using different views of ShapeNet \cite{shapenet} car scenes, and evaluate the metric learning property of the embeddings produced by different models: a model induces a useful representation if the embeddings of different views of the same scene are mapped closer in vector space compared to views of different scenes.
% \newpage % remove this?

\begin{samepage}
\textbf{Main contributions.}
Our main contributions are summarized below:
\begin{itemize}
    \item We constructed a dataset derived from ShapeNet \cite{shapenet} containing renders of 3D models, which is designed to explore 3-D awareness in metric learning via scene retrieval.
    \item Performed zero-shot retrieval using embeddings from pre-trained image encoders, and showed that latent embeddings from 3D diffusion outperform those of classification CNNs.
    \item Demonstrated that contrastive learning improves scene retrieval and learns an embedding with (imperfect) view invariance.
    \item Report ablations showing dynamics of varying number of reference images per scene as well as number of images seen during training.
\end{itemize}
\end{samepage}

\section{Related Work} \label{section_related_work}
\subsection{Neural rendering for discriminative tasks}
We note that the space of using generative (e.g. diffusion) approaches to solve discriminative tasks is remarkably underexplored. Mainly, a handful of works propose to use 2D diffusion models to solve specific dense prediction, learning to denoise intermediate artifacts like bounding boxes, depth maps, and filters by reversing a manually constructed noising process \cite{diff_generalist, diff_depth, diff_inst, diff_ensemble, diff_med}.

Notably, there have been no works analyzing \textit{3D neural rendering} models for discriminative tasks, and simultaneously no works directly analyzing using the strong visual priors of generative models using lens of representation learning. We aim to provide preliminary explorations on both of these areas, investigating a 3D diffusion model trained on a large dataset of 3D scenes, and analyzing one measurable aspect of the resultant 3D-aware priors — view independence. Our results encourage further exploration of this field and show that 3D diffusion models indeed induce useful representations zero-shot.

\subsection{Metric learning for image retrieval} \label{section_metric_learning}
The state-of-the-art on 2D metric learning retrieval benchmarks is dominated by classic convolutional models trained on the metric learning objective using increasingly sophisticated contrastive loss functions. After the advent of triplet loss \cite{triplet_loss}, which explicitly penalizes the distance between embeddings of same-labeled examples and rewards the distance (up to a threshold) between embeddings of different-labeled examples, various improvements were proposed that improve separability, including adding regularization terms relating to clustered classes tightly around class centers \cite{center_loss}, normalizing the distances between such centers \cite{sphereface}, and enforcing the decision margin in a more natural way (i.e. in angle space) \cite{arcface}. Since we focus on preliminary investigations on the choice of the pre-trained backbone model rather than the choice of subsequent contrastive training, we use the simple SimCLR contrastive training setup with cross-entropy loss (which was determined in the paper to outperform triplet loss), with details in Section \ref{section_simclr}.

\subsection{EfficientNet}
EfficientNet \cite{efficientnet} is a popular convolutional neural network baseline for a variety of scene understanding tasks, including achieving the state-of-the-art on transfer learning datasets such as CIFAR-100 \cite{cifar100} and ImageNet-1K \cite{imagenet}. It is also commonly used as a backbone for the metric learning models described in Section \ref{section_metric_learning}. Its main contribution was decreasing the number of parameters by an order of magnitude while producing significantly improved performance using observations about optimal scaling laws for CNNs (i.e., finding the relationship between scaling width, depth, and resolution). Here, we use the second-to-last layer of the network as the canonical latent representation for discriminative tasks.

\subsection{Shap-E}
Shap-E \cite{shape} is a state-of-the-art conditional generative model that generates 3D assets conditioned on 2D image or text by learning implicit function representations. Specifically, the work trains an encoder that maps 3D assets into latent codes representing the parameters of an implicit function, then trains a conditional diffusion model on these latent codes — we use the learned latent codes as the natural learned representation for image and scene understanding tasks.

\subsection{SimCLR} \label{section_simclr}
To understand the effect of self-supervised training on the retrieval database (i.e., learning using the supervision that different views of the same database image belong to the same scene), we apply the SimCLR framework \cite{simclr} for contrastive learning. SimCLR proposed a simple method for minimizing contrastive loss in the latent space with the following steps: \cite{lil_simclr}
\begin{enumerate}
    \item Sample a random data augmentation $t$ (e.g. random flip, crop, resize) to produce pairs of images that represent different views the same image, $x, x' = t(x)$, for a given batch.
    \item Compute the embedding of each (possibly transformed) image by passing it through an encoder and a projection head.
    \item Treating $(x,x')$ as a positive pair and all other combinatorial pairs of (possibly transformed) images within the batch as in-batch negatives, minimize the contrastive loss over cosine similarity $\simm(\cdot, \cdot)$.
    \[
        \mathcal{L}^{(i,j)} = -\log \frac{\exp(\simm(z_i,z_j)/\tau)}{\sum_k \ind{k \ne i} \exp(\simm(z_i,z_k)/\tau}
    \]
    where $\tau$ is a temperature parameter.
\end{enumerate}
The final representations are then the encoder functions (without the projection head), which can then be used to compute representation vectors for possibly held-out images on downstream tasks. In our setup, we treat different views of the same scene as the data augmentation in SimCLR, since we wish our embedding to cluster these views together.

\section{Methods}
\subsection{Retrieval task}
To understand whether the latent space of the Shap-E \cite{shape} neural rendering model better captures 3D-aware properties compared to that of classical convolutional models like EfficientNet \cite{efficientnet}, we can evaluate the view independence of each encoder via the metric learning property, where views of the same image should be mapped to vectors with high similarity. Formally, we can measure this for each encoder $f$ by computing its embedding for each input view of each scene $x_i^j$ (indicating this is view $j$ of scene $i$), where we would like it to be the case that
\[
    \simm(f(x_{i_1}^{j_1}), f(x_{i_1}^{j_2})) \le \simm(f(x_{i_2}^{j_3}), f(x_{i_3}^{j_4}))
\]
for all $i_2 \ne i_3$ and all $i_1,j_1,j_2,j_3,j_4$. (i.e., any pair of same-scene views should be closer than any pair of views from different scenes) Here, $\simm(\cdot,\cdot)$ is a similarity function such as cosine similarity.

\begin{figure}
\begin{centering}
\includegraphics[width=0.5\textwidth]{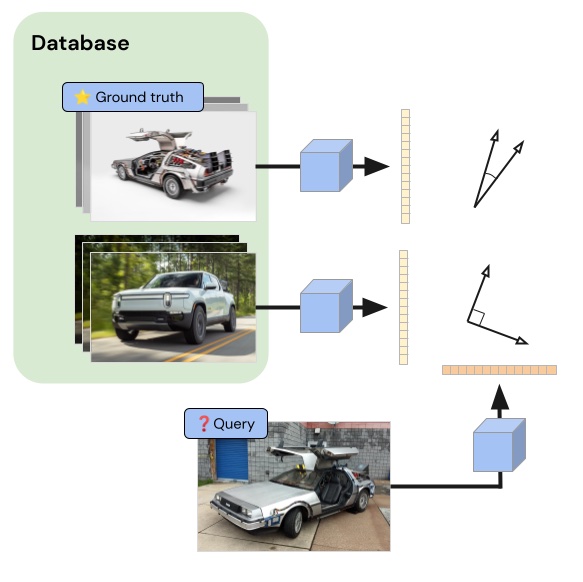}
\caption{\label{fig_retrieval} Setup of the metric learning retrieval task}
\end{centering}
\end{figure}

We can measure this by constructing the canonical 2D image retrieval task, where we construct a \textit{database} of $n$ different scenes, with $k_{db}$ different image views each, and an \textit{query set} of $m \le n$ of those scenes with $k_{query}$ different image views each. The retrieval task is as follows: given a query image, return the most relevant scene from the database. The formulation that evaluates metric learning properties imposes the following retrieval algorithm:
\begin{enumerate}
    \item Compute $f(q)$ for the query image $q$.
    \item Compute $f(x_i^j)$ for each view of each database scene.
    \item Compute the similarities $\simm(f(q),f(x_i^j))$ for all $i,j$ according to similarity function $\simm(\cdot,\cdot)$, and return the top retrieved image
    \[
    (i,j) = \argmax_{i,j} \simm(f(q),f(x_i^j))
    \]
    If we want the top $k$ images, this corresponds to retrieved images
    \[
    (i_l,j_l)_{l=1,...,k} = \argmax_{(i_l,j_l)_{l=1,...,k}} \simm(f(q),f(x_{i_l}^{j_l}))
    \]
    (note that some images can be of the same scene).
\end{enumerate}
This is depicted graphically in Figure \ref{fig_retrieval}.

In particular, the second — most costly — step can be precomputed in a single indexing step instead of before each new query image, and the third step can be implemented via highly optimized Maximum Inner Product Search algorithms, which makes this method highly efficient. This is notably the state-of-the-art algorithm for retrieval pipelines for dense retrieval methods across many domains, e.g. for text document retrievals. Thus to evaluate the view independence of the representations of different models, we substitute them in as different retrievers $f$ using this view-based dataset construction, and evaluate the resulting retrieval performance using the following standard metrics averaged over queries:
\begin{enumerate}
    \item \textbf{Mean Reciprocal Rank (MRR)}@$k$: among the top $k$ retrieved images, if none are from the ground truth scene, return 0. Else return $1/k'$, where $k'$ is the smallest index of a candidate that is from the ground truth scene. In this paper, unless otherwise specified, we choose $k=\infty$ (i.e. we find the rank over all predictions) and simply denote this as MRR.
    \item \textbf{Recall}@$k$: among the top $k$ retrieved images, return the fraction of them that are from the ground truth scene.
\end{enumerate}

Importantly, the query and database image views for each scene are disjoint in task construction, and in the \textit{zero-shot} setting neither set is available at training time, mirroring real-world applications where users can upload images tagged with the scene they are related to (e.g., the entity name of a global landmark), as well as input queries where they are interested in retrieving the scene for a given new image (e.g., a new picture they took of some unknown global landmark in the database). Rather, the encoder $f$ must learn image understanding (and in particular 3D-aware ones such as view independence) by being trained on other datasets. Note that unfortunately, the exact Shap-E dataset is unreleased, so it is possible the scenes from our retrieval dataset comprises a small part of the their training data during the generative training process, but we specifically refrain from applying any discriminative training on any of the retrieval encoders during zero-shot evaluation.

\textbf{Implementation.} We used the validation set of the cars single-category subset of the ShapeNet dataset \cite{shapenet} ($m = n = 300$ scenes), and unless otherwise stated used $k_{db}=20,k_{query}=1$, mimicking the real-life lack of more views available in the database but typically only one per query. We extracted latent spaces from both models as detailed in Section \ref{section_related_work}, using a PyTorch reimplementation of SimCLR \cite{simclr_pytorch}. We use cosine similarity in all experiments.

\subsection{View aggregation}
Some retrieval settings enforce the use of a single vector representation per scene (e.g. at large scales, for computational efficiency). Thus we also examine whether the latent representations of neural-rendering-based and classical models have other semantic properties, such as the ability to be aggregated under a mean. To evaluate this, we run a modified retrieval pipeline where the vector representation of a scene is the elementwise mean of the vector representations of each image view of that scene available in the database:
\[
f(i) = \frac{1}{k_{db}} \sum_j f(x_i^j)
\]
and then compare the resulting \textit{scene} representations using the similiarity function to get top $k$ scenes. We denote this as the \textit{mean-aggregated setting}.

\subsection{Training on the database images}
Alongside approaching the task zero-shot, we are also interested in how well the latent spaces adapt to the evaluation database, i.e., if models are given training-time access to the database images. Notably, there is still no training-time access to the query images. This emulates the real-world scenario where a large database of scenes of interest is uploaded (e.g., famous world landmarks), and an indexing service can perform further training on the encoder using the fixed database before deployment, although the potential query images users are interested in are arbitrary.

To do so, we trained the latent embeddings of Shap-E contrastively using the SimCLR \cite{simclr} setup (details in Section \ref{section_related_work}) using randomly sampled views of the same scene as positive pairs and using in-batch negatives.

\textbf{Implementation.} We froze the encoders $f$ and trained a multi-layer (3 layers for EfficientNet and 5 for Shap-E performed best) fully-connected MLP as well as 1-layer projection head MLP on top using the contrastive objective, Following the original SimCLR paper, we discarded the 1-layer projection head during evaluation and treated the layer beneath as the representation. The depth and width were chosen using a hyperparameter sweep.

\section{Results}
\subsection{Image Retrieval}
\begin{figure*}
    \centering
\begin{tabular}{lccc} \toprule
   & \multicolumn{1}{p{2.5cm}}{\centering Recall@1}
   & \multicolumn{1}{p{2.5cm}}{\centering Recall@5}
   & \multicolumn{1}{p{2.5cm}}{\centering MRR} \\
   \midrule
   w/o learned encoder\\
   \quad EfficientNet & 0.82 & 0.86 & 0.85\\
   \quad Shap-E & \textbf{0.86} & \textbf{0.96} & \textbf{0.90}\\ \midrule
   w/ learned encoder\\
   \quad EfficientNet & \textbf{0.94} & \textbf{1.00} & \textbf{0.96}\\
   \quad Shap-E & 0.88 & 0.94 & 0.91\\
   \bottomrule
\end{tabular}
\caption{Metric learning retrieval results.} \label{table_daresults}
\end{figure*}

\begin{figure*}
    \centering
\begin{tabular}{lccc} \toprule
   & \multicolumn{1}{p{2.5cm}}{\centering Recall@1}
   & \multicolumn{1}{p{2.5cm}}{\centering Recall@5}
   & \multicolumn{1}{p{2.5cm}}{\centering MRR} \\
   \midrule
   w/o learned encoder\\
   \quad EfficientNet & \textbf{0.62} & \textbf{0.76} & \textbf{0.70}\\
   \quad Shap-E & 0.42 & 0.74 & 0.56\\\midrule
   w/ learned encoder\\
   \quad EfficientNet & \textbf{0.88} & \textbf{0.98} & \textbf{0.92}\\
   \quad Shap-E & 0.82 & \textbf{0.98} & 0.88\\
   \bottomrule
\end{tabular}
\caption{Metric learning retrieval results, with a database of single-scene mean-aggregated representations.} \label{table_mean_metric}
\end{figure*}

\begin{figure*}
\begin{centering}
\includegraphics[width=0.6\textwidth]{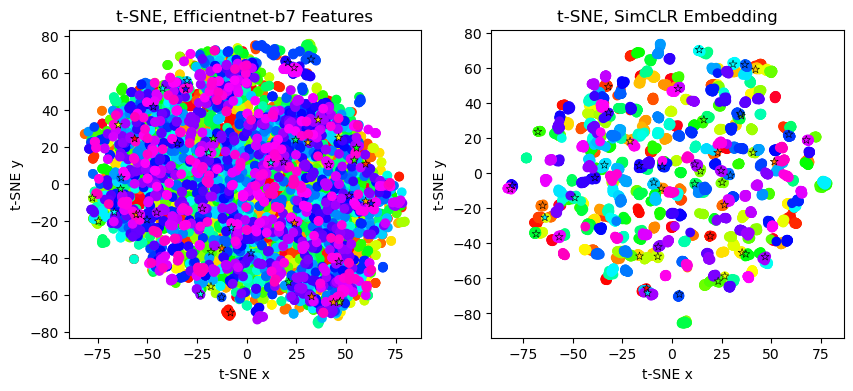}
\includegraphics[width=0.6\textwidth]{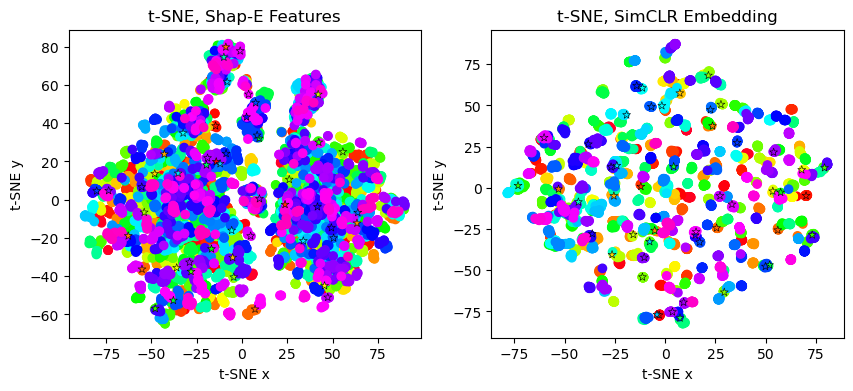}
\caption{\label{fig_tsne} t-SNE plots of EfficientNet (top) and Shap-E (bottom) representations before (left) and after (right) contrastive training. Dots represent database vectors and stars represent query vectors, with color denoting the scene.}
\end{centering}
\end{figure*}

\begin{figure*}
\begin{centering}
\includegraphics[width=0.6\textwidth]{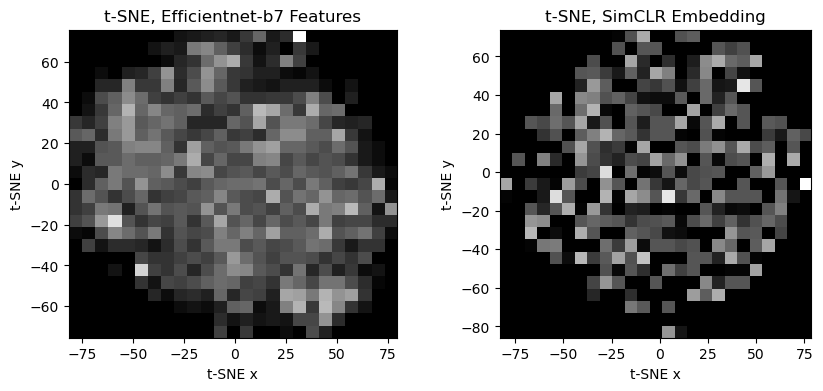}
\includegraphics[width=0.6\textwidth]{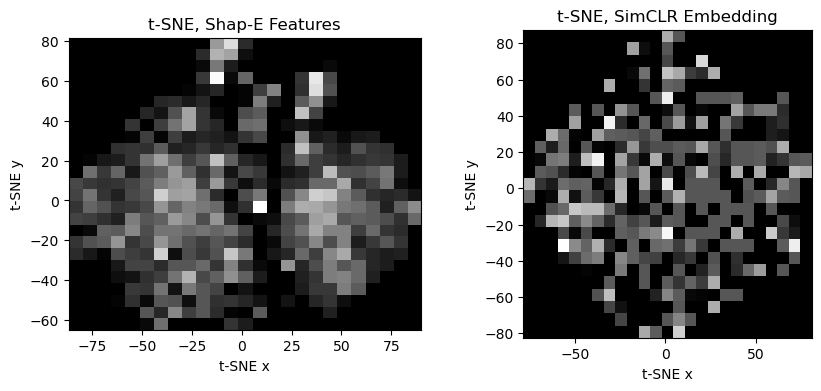}
\caption{\label{fig_tsne_heatmap} Heatmaps of binned t-SNE plots of EfficientNet (top) and Shap-E (bottom) representations before (left) and after (right) contrastive training.}
\end{centering}
\end{figure*}

We report metric learning retrieval results in Table \ref{table_daresults}, including the performance of both the zero-shot raw and contrastively trained EfficientNet and Shap-E embeddings. We visualize the effect of training by showing t-SNE plots of EfficientNet and Shap-E embeddings before and after training in Figure \ref{fig_tsne} as well as report just the density of those t-SNE plots to more cleanly visualize the separability of the clusters of images corresponding to different scenes in Figure \ref{fig_tsne_heatmap}. For these results, we used $k_{db}=20, k_{query}=1$ and trained on the same 20 images.

Overall, we find that Shap-E embeddings substantially outperform the EfficientNet embeddings zero-shot on all retrieval metrics, although SimCLR training improves both representations, including making EfficientNet embeddings competitive with the Shap-E embeddings. This is corroborated by the t-SNE visualizations, which show the Shap-E embeddings to be slightly more separated, and the trained versions of both embeddings to be much more separated, with the query embeddings almost always being mapped on top of a cluster of database embeddings (visualized as each star symbol being mapped on top of the circle symbols of the same color). The density t-SNE plots also support this, with the trained embeddings clearly better separated and sparser with higher peaks than their untrained counterparts.

We conjecture that the Shap-E latent space is particularly well-suited to cosine similarity because it was trained to reconstruct 3D meshes from input images; similar images should be encoded in similar feature vectors. EfficientNet, however, is trained to classify images in the diverse ImageNet \cite{imagenet} dataset. The feature space learned by EfficientNet's penultimate layer must be separable by the final fully connected layer, but is not necessarily endowed with a similarity metric. However, this metric property is explicitly optimized by SimCLR, which is likely what allows both sets of learned embeddings to be strong. 

\subsection{Mean Aggregation}
We report metric learning retrieval results in the mean-aggregated setting in Table \ref{table_mean_metric}. For this result, we used the same properties as in our main results ($k_{db} = 20, k_{query} = 1$), but take represent each scene by taking the mean over $k_{db}$ representative vectors. Overall, we find that mean-aggregation degrades the performance of zero-shot retrieval on both the Shap-E and Efficient embeddings. However, contrastive learning dramatically improves the performance of mean-aggregation, and the mean-aggregated retrieval results after training are competitive with ordinary retrieval on the trained embeddings in Table \ref{table_daresults}. This suggests that by pre-processing our database with contrastive learning and per-scene mean-aggregation, we can speed up similarity search on future queries without sacrificing recall.

\section{Analysis}
We examine the effect of changing the number of views per scene available in the database at query time, $k_{db}$, as well as the number of views per scene available as positive pairs in SimCLR training, which we will denote $k_{train}$.

\subsection{Effect of $k_{db}$ on retrieval}
\begin{figure*}
\begin{centering}
\includegraphics[width=0.9\textwidth]{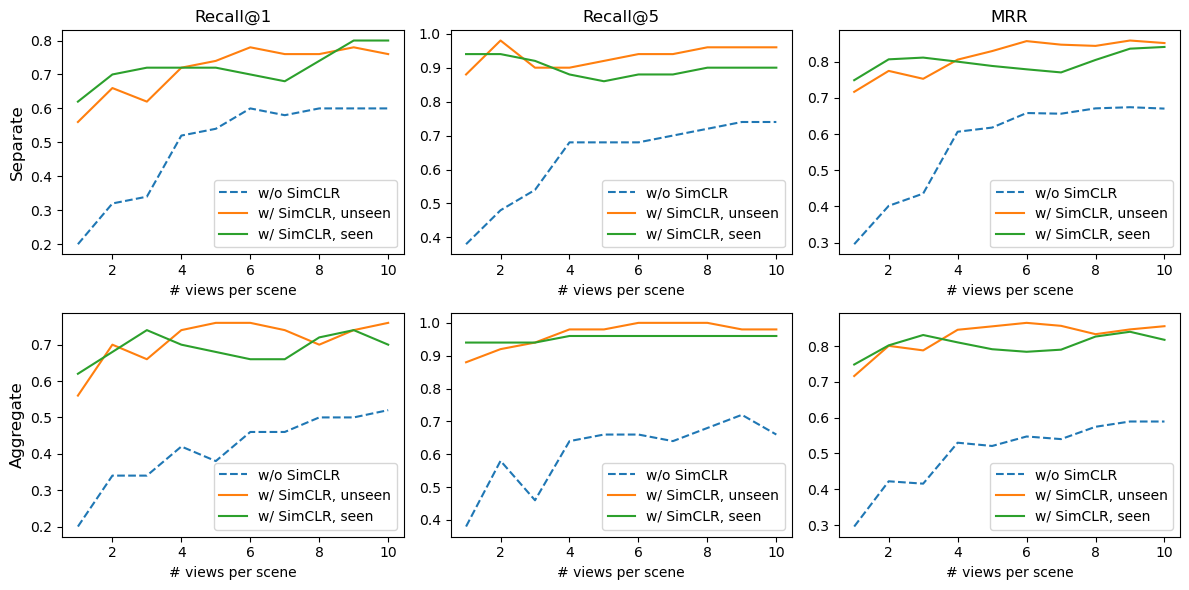}
\caption{\label{fig_effnet_k_ablation} Effect of changing $k_{db}$ with (bottom) and without (top) aggregation on metric learning retrieval performance, for EfficientNet encoder. The dotted line refers to zero-shot (i.e., untrained) performance, and the two solid lines refer to trained performance where the database images at evaluation are seen or unseen during training, respectively. Query images are always unseen.}
\end{centering}
\end{figure*}

\begin{figure*}
\begin{centering}
\includegraphics[width=0.9\textwidth]{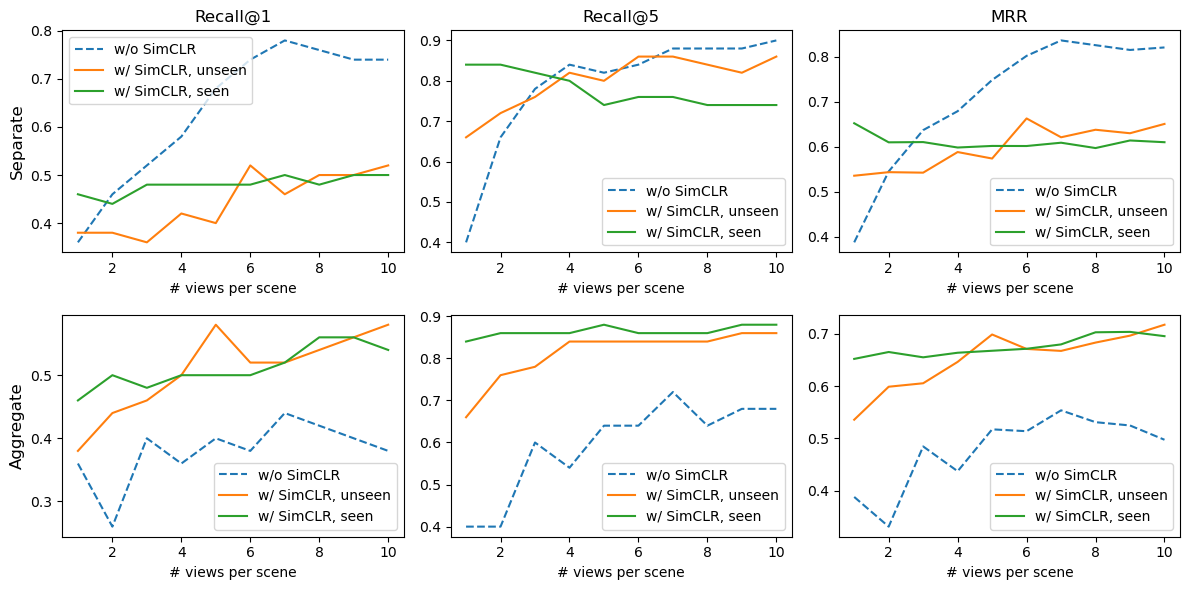}
\caption{\label{fig_shape_k_ablation} Effect of changing $k_{db}$ with (bottom) and without (top) aggregation on metric learning retrieval performance, for Shap-E encoder. Same formatting details as Figure \ref{fig_effnet_k_ablation}.}
\end{centering}
\end{figure*}

We plot evaluation results on databases with different values of $k_{db}$ for both trained and untrained (using a dotted line) EfficientNet embeddings, in both the standard and mean-aggregated retrieval settings in Figure \ref{fig_effnet_k_ablation}, and the same for Shap-E embeddings in Figure \ref{fig_shape_k_ablation}. For these experiments, we use a fixed $k_{train}=10$ images per scene in training.

Note that in order to evaluate the robustness of the SimCLR-trained embeddings, alongside evaluating with a database of images including some seen at training time, we also consider the alternative setting where we evaluate on a disjoint database of \textit{unseen} images. In the latter case, the training images, database images, and query image are \textit{all mutually disjoint} for every scene. In this scenario, we only see good results if the training process learns general view-independent properties of the scene, instead of memorizing the specific positive pair images. Since we in fact find that the unseen and seen performances are comparable, we conclude the SimCLR contrastive learning process learns to encode view independence robustly.

\textbf{Standard (un-aggregated) setting.}
We find that more views improves zero-shot embedding performance for both kinds of representations, with major improvements up to about 4 views for EfficientNet and higher (about 8) for Shap-E. One interpretation of this is that view independence has not fully been captured by either encoder, as seeing more views of a scene adds a large amount of information compared to what would be expected if the model fully extrapolated from views seen and only learned information from new views about previously occluded, asymmetric parts of the car (since the cars are highly symmetric this is expected to be small).

In contrast, training mostly removes this $k_{db}$ dependence when encoding both seen and unseen database images, which points to SimCLR being an effective training method for instilling view independence. 

\textbf{Mean-aggregated setting.}
Zero-shot, we find that both encoders admit some ability to represent scenes using mean aggregation on representations of individual views, as performance improves with more views being aggregated, pointing to new information being captured to some degree. Unfortunately, the raw performance under multiple retrieval metrics for both EfficientNet and Shap-E is poor.

However, we find that training consistently improves performance for both encoders in this setting. In accordance with our previous findings in the standard setting, the trained encoders have little dependence on $k_{db}$ in the aggregate setting. This reinforces the robustness of training, but also points to the fact that we may not be able to draw strong conclusions about whether the trained embeddings admit aggregation, since they already encode some degree of view independence and do not significantly benefit from more views — for example, in the ideal case where training endows full view independence (which we do not achieve here), aggregation would be between identical embeddings and thus be trivial, so some version of this effect is expected. It suffices to note that since both encoders receive a boost in base performance on a single view when trained and aggregating it with more images does not hurt performance for EfficientNet and slightly improves it for Shap-E, the training process slightly promotes the ability for representations to be aggregated. This is itself a positive result, since aggregation leads to significant computational and efficiency gains for downstream retrieval applications, so we conclude that the training process can significantly improve the EfficientNet and Shap-E representations (via improving base performance) while also preserving those practical gains (when aggregating additional image representations).

\subsection{Effect of $k_{train}$ on retrieval}
We evaluate metric learning retrieval performance for trained EfficientNet and Shap-E embeddings using different values of $k_{train}$. We again consider unseen and seen settings for the trained encoders, and include the zero-shot, untrained performance of both encoders as a baseline, shown as a dotted line. For these experiments, we fix $k_{db} = 4$.

We see that training with more views indeed intuitively leads to substantially better performance for EfficientNet embeddings. In contrast, there is no clear pattern for Shap-E. We see some instability in training runs for intermediate values of $k_{train}$, and for very low values of $k_{train}$, the trained Shap-E actually often does worse at the metric learning retrieval task than its zero-shot counterpart. This may imply that it is hard to learn view independence in a many-layer MLP head on top of a high-dimensional Shap-E model with only a few views per scene, and in particular that the optimization landscape for training on top of Shap-E is less amenable to learning view independence than the one for training on top of EfficientNet, despite the fact that the zero-shot performance of Shap-E is higher.

Again, we find that unseen and seen performance across the board is about the same, pointing to SimCLR training being robust.

\section{Discussion}
The main limitation of this work is that it focuses on a single domain (synthetic Blender cars from ShapeNet \cite{shapenet}) and one neural-rendering-based encoders, namely Shap-E \cite{shape}. We find that this encoder exhibits useful properties as a representation learner, opening up a new area for research, but the result would be stronger if it were replicated across multiple different encoders, especially given the diversity of different neural rendering approaches, including non-diffusion-based ones. Replicating this across multiple large benchmarks and a high-quality in-the-wild dataset would also significantly strengthen the results Under ideal circumstances, we would also want access to the details of the training set of Shap-E to make sure it is definitively disjoint from our retrieval task.

Another limitation of this work is the large memory footprint of the proposed vector embeddings. Each Shap-E embedding is a $1,048,576$-dimensional vector, requiring $4.8$MB per embedding or $96$MB per scene (with $k_{db} = 20$) to store in memory. For large retrieval databases, this memory consumption may be too costly to achieve zero-shot Shap-E retrieval at scale. After training, however, each learned embedding takes on the order of kilobytes to store, suggesting that such a database may scale after pre-processing with SimCLR and mean-aggregation.

\section{Conclusion}
We conclude that neural-rendering-based generative models induce useful representations in latent space, encoding images in a way that captures view independence as well as the ability to aggregate views to form a single coherent scene vector, and performing quantitatively better on these abilities on various retrieval metrics than classical trained on discriminative tasks in the zero-shot setting. We find that the primary gains are powerful zero-shot priors, and that rendering-based and classic models both perform well once the model is given access to database images at training time. As well as achieving improved performance on 2D image retrieval, this opens up new possibilities for non-generative tasks, including 3D decision-making, recognition, and visual question-answering, and suggests that a promising direction forward in those areas involves adapting the priors of existing state-of-the-art generative models.

\section{Acknowledgements}
The authors would like thank Fangyin Wei, Professor Felix Heide, Evan Dogariu, and other students in COS526 for helpful discussions and kind support.

\todo{expand related work and add 10 citations}
\todo{clean up and make readme / entry point for source code and submit}

{\small
\bibliographystyle{ieee_fullname}
\bibliography{egbib}
}

\begin{figure*}
\begin{centering}
\includegraphics[width=0.9\textwidth]{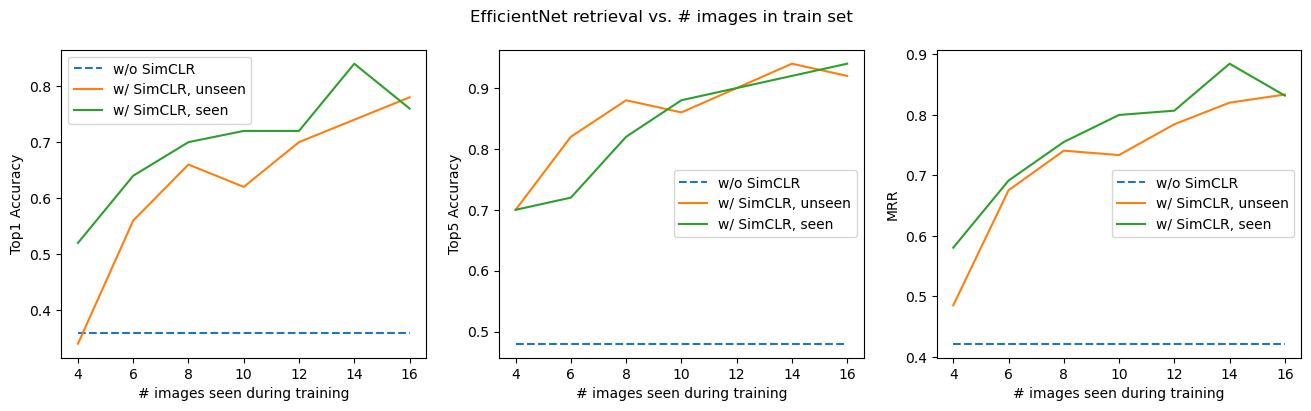}
\includegraphics[width=0.9\textwidth]{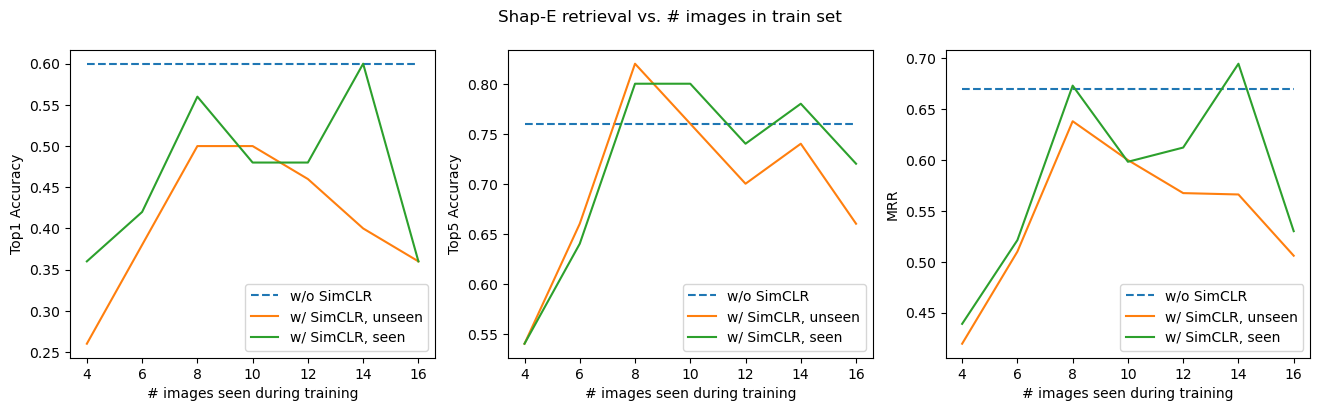}
\caption{\label{fig_effnet_train_k_ablation} Effect of changing $k_{train}$ on metric learning retrieval performance, for EfficientNet (top) and Shap-E (bottom) encoders. The dotted line refers to zero-shot (i.e., untrained) performance, and the two solid lines refer to trained performance where the database images at evaluation are seen or unseen during training, respectively. Query images are always unseen.}
\end{centering}
\end{figure*}

\end{document}